# A Stochastic Approach to STDP

Runchun Wang, Chetan Singh Thakur, Tara Julia Hamilton, Jonathan Tapson, André van Schaik
The MARCS Institute, Western Sydney University, Sydney, NSW, Australia
mark.wang@westernsydney.edu.au

*Abstract*— We present a digital implementation of the Spike Timing Dependent Plasticity (STDP) learning rule. The proposed digital implementation consists of an exponential decay generator array and a STDP adaptor array. On the arrival of a pre- and post-synaptic spike, the STDP adaptor will send a digital spike to the decay generator. The decay generator will then generate an exponential decay, which will be used by the STDP adaptor to perform the weight adaption. The exponential decay, which is computational expensive, is efficiently implemented by using a novel stochastic approach, which we analyse and characterise here. We use a time multiplexing approach to achieve 8192 (8k) virtual STDP adaptors and decay generators with only one physical implementation of each. We have validated our stochastic STDP approach with measurement results of a balanced excitation/inhibition experiment. Our stochastic approach is ideal for implementing the STDP learning rule in large-scale spiking neural networks running in real time.

## I. Background

The Spike Timing Dependent Plasticity (STDP) algorithm [1], which has been observed in the mammalian brain, modulates the weight of a synapse based on the relative timing of pre-synaptic and post-synaptic spikes. In STDP, the synaptic weight will be increased (or decreased) if a pre-synaptic spike arrives several milliseconds before (or after) the post-synaptic spike fires. This learning rule is computationally intensive as it exponential functions and divisions.

In neuromorphic systems, various implementations of the STDP algorithm have been proposed, such as a circuit based on analogue blocks and flip-flops [2], a bistable synapse with a very compact analogue implementation of the STDP [3], and analogue blocks and switches to implement exponential STDP [4]. We have previously presented a compact implementation of the STDP using linear decays [5], [6]. Here, we present a novel stochastic approach that works with our previous system and can efficiently implement the STDP operations.

## II. Exponential Decay

### A. Infinite Impulse Response (IIR) filter approach

A discrete time first order exponential decay implemented with an IIR filter can be expressed by the following equation:

$$V[t+1] = \alpha V[t] \quad (1)$$

where, $t$ represents the index of the time step, and $V[t]$ represent the previous value of $V$ and the IIR filter constant $\alpha$ is defined as:

$$\alpha = \frac{\tau}{\tau+1} \quad (2)$$

where, $\tau$ is the time constant (in clock cycles) and the decay $d$ is given by:

$$d = V[t] - V[t+1] = \frac{V[t]}{\tau+1} \quad (3)$$

When $\tau$ is large, $\alpha$ is only a little less than 1, and a large number of bits are needed to encode its value accurately in a digital system. If the number of bits used to encode $V$ is less than, the number of bits used to encode $\alpha$, the above recursive multiplication just results in a linear decay.

This situation occurs, for example, when simulating a neural network with many millions of neurons using time multiplexing [7]–[10]. With a standard IIR filter approach, a large number of bits would be needed for each state variable to achieve enough resolution to calculate long time constants. Large memory storage per state variable will result in a communication bottleneck, since only a few bits can be exchanged with the memory in a single clock cycle.

### B. Stochastic decay

Instead of implementing the IIR multiplication directly, we can instead multiply $V$, encoded with much fewer bits than $\alpha$, by the IIR factor $\alpha$ and then add a random number $r$ to the multiplication result. Mathematically, the method can be written as:

$$V[t+1] = int(\frac{\tau}{\tau+1}V[t] + r[t]) \quad (4)$$

where, $r[t]$ is a random number drawn from a uniform distribution in the range (0,1). This is effectively a form of dithering to deal with the rounding of $V$ to an integer value. For example, in our implementation discussed below we use 4 bits for $V$, 7 bits for $r$, and 9 bits for $\alpha$.

This work has been supported by the Australian Research Council Grant DP140103001. This work was inspired by the Capo Caccia Cognitive Neuromorphic Engineering Workshop 2014 and Telluride Neuromorphic workshop 2015.

The decay is then given by:

$$d = V[t] - V[t+1] \quad (5)$$

$$= V[t] - int(\frac{\tau}{\tau+1}V[t] + r[t]) \quad (6)$$

$$= int\left(\frac{V[t]}{\tau+1}\right) + X[t] \quad (7)$$

$$p = P(X[t] = 1) = \frac{V[t]}{\tau+1}\%1 \quad (8)$$

where $X[t]$ is a random binary variable, $p$ is the probability that $X[t] = 1$, % is the modulo operation and $int\ (V[t]/(\tau+1))$ is the integer part of $V[t]/(\tau+1)$. The expected value of $X$ is given by (8), and simply represents the fractional part of $V[t]/(\tau+1)$. The expected value for the decay is thus the integer plus fractional part of $V[t]/(\tau+1)$ and thus equal to the IIR decay in (3), but we now only need to store a few bits for $V[t]$.

*C. Characterisation of variance*

Our stochastic approach not only reduces the storage needed, but also introduces variability between the STDP synapses, even when they time multiplex the exact same digital synapse. This variability makes the networks more realistic simulations of biological neural networks. Other digital implementations typically need to provide explicit sources of randomness when simulating neural networks.

The stochastic part of the decay is fully determined by $X[t]$, which is either 0 or 1. Thus the number of time steps $n$ it takes to get a single stochastic decrement is given by the geometric distribution:

$$P(n) = (1-p)^{n-1}p \quad (9)$$

The variance for this distribution is given by:

$$Var(n) = \frac{1-p}{p^2} = \frac{(\tau+1)^2}{V^2} - \frac{\tau+1}{V} \quad (10)$$

The variance is thus very large when $\tau$ is much larger than $V$. In (4), $r$ is drawn from a uniform distribution in the range (0,1). Reducing the variance can be effectively achieved by limiting $r$ in a smaller range as long as the following condition is met:

$$\alpha V + min(r) < V \quad (11)$$

Otherwise $V$ will not decay. It is obvious that this condition is most critical when $V$ is 1. For digital implementations, the most efficient way to generate random numbers is to use linear feedback shift registers (LFSRs), which do not have the value 0 as a possible output. Thus we can express this condition (with $V = 1$) as:

$$\frac{\tau}{\tau+1} + min(r) = \frac{\tau}{\tau+1} + \frac{1}{2^L} < 1 \quad (12)$$

$$\tau < 2^L - 1 \quad (13)$$

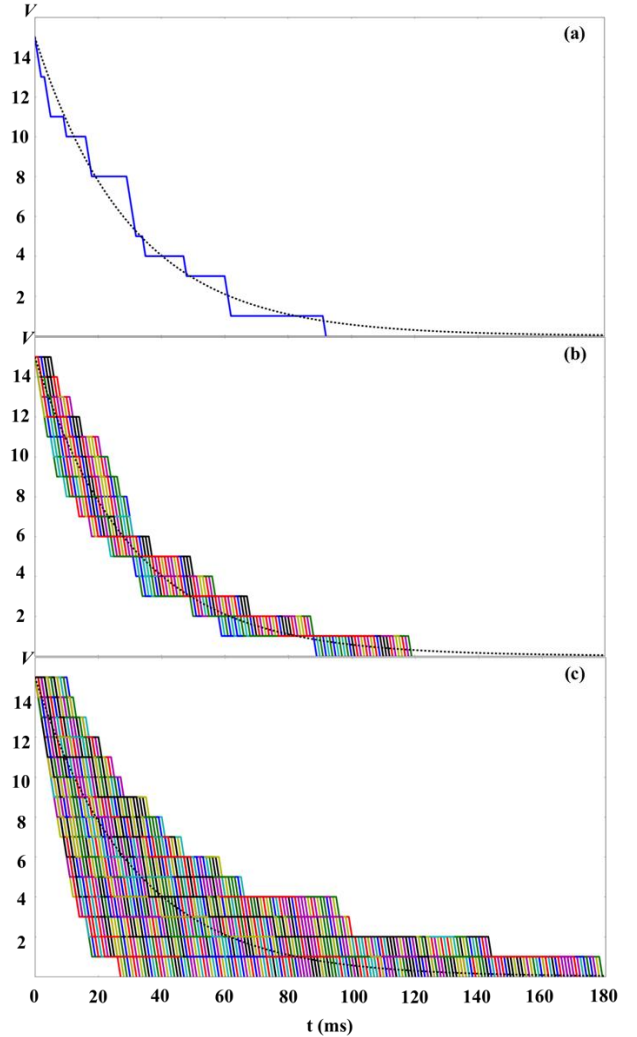

**Fig. 1. Exponential decay obtained by using the stochastic approach.** Here we use a 1 ms clock cycle. The dashed line is the IIR decay trace with a time constant $\tau$ of 30 ms ($\alpha = 495/512$, a 9-bit number). $V$ is stored as a 4-bit integer with an initial value of 15. (a) An example exponential achieved when using a 5-bit LFSR; (b) All possible decays with a 5-bit LFSR with different seeds; and (c) Exponential decays achieved using a 9-bit LFSR and 1023 different random seeds. It is clear that the variance of the exponential decays achieved with the 9-bit LFSR is much larger than that with the 5-bit LFSR.

where, $L$ is the length of the LFSR. For example, the maximum time constant that a 5-bit LFSR can achieve is 30 clock cycles. Using a 9-bit LFSR for the same time constant will create much larger variances (see Fig. 1). Hence the principle to reduce the variances is to use the LFSR with the minimum length that can still achieve the time constant.

## III. HARDWARE IMPLEMENTATION

### A. Learning rule

In our hardware implementation, the amount of synaptic modification is summarised by the following standard exponential STDP equations:

$$\Delta w = \begin{cases} A^+ exp(\Delta t/\tau_+), & if\ \Delta t < 0 \\ -A^- exp(\Delta t/\tau_-), & if\ \Delta t \geq 0 \end{cases} \quad (14)$$

where, $\Delta w$ is the modification of the synaptic weight, $\Delta t$ is the arrival time difference between the pre- and post-synaptic spike. $A^+$ and $A^-$ determine the maximum amounts of synaptic modification for each spike pair. $\tau_+$ and $\tau_-$ are the time constants and control the rate of decay for potentiation and depression portions of the curve, respectively. As we focus on the low-cost hardware implementation of the exponential-type STDP, quantifying the effects of the STDP learning rules on the synaptic weights [11] are outside the scope of this paper. In the work reported here, we use $\tau_+ = \tau_- = 20$ ms and $A^+ = A^- = 1$ throughout. Hence, the $\Delta w$ is simply $V[t]$ in equation (4).

### B. Topology

In our previous work [5], [6], we implemented a time multiplexed (TM) synaptic plasticity adaptor array that is separate from the neurons in the neural network. Each adaptor (in that array) performs synaptic plasticity, (such as STDP), according to the arrival times of the pre- and post-synaptic spikes assigned to it and sends out the updated weight to the post-synaptic neuron in the neural network. Since this strategy provides great flexibility for building complex large-scale neural networks, we chose to use the exact same architecture as in [6] to implement an exponential-type STDP adaptor array (see Fig. 2). It consists of a controller, a Master RAM, a TM STDP adaptor array and a TM exp-decay generator array, all of which, with the exception of the last one, are identical to ones presented in [6]. The TM adaptor array and the TM exp-decay generator array are both configured to have 8192 (8k) units, each TM exp-decay generator being assigned to one TM STDP adaptor. Thus, the TM time window generator array in [6], which generates a linear decay, is replaced by the exp-decay generator array in the work presented here.

The exponential STDP adaptor array operates in the exact same manner as the digital synaptic adaptor array in [6]. The controller receives pre- and post-synaptic spikes from the neuron array and assigns them to the corresponding TM STDP adaptors according to their addresses. Each TM exp-decay generator will start an exponential decay when either a pre- or post-synaptic spike arrives, which will be used by the corresponding TM STDP adaptor to determine the weight update. As we assume that the adaption will not be carried out if the pre- and post- synaptic spikes arrive simultaneously, thus only one TM exp-decay generator will be needed. The STDP adaptor will carry out the weight adaption using its output $V_{[t]}$. The weight values are stored in the local cache and the Master RAM. The stored weight values will also be sent out to the corresponding neuron in the neural network for the post-synaptic current generation.

### C. Time-multiplexed exponential decay generator array

The decay generator array was implemented by using time multiplexing to achieve 8k TM exp-decay generators using only one physical exp-decay generator. The global counter processes each TM exp-decay generator sequentially. Each TM exp-decay generator uses a time slot of 25 clock cycles (125 ns with 200 MHz clock frequency) to complete its processing to maintain an update rate of 1 kHz (the corresponding time step is about 125 ns×8k=1 ms).

In each time slot, the global counter will read the value of the $V[t]$ (a 4-bit integer) from the Decay RAM with a size of 8k×4bit. $V[t]$ will be reset to $V_{init}$ (set to 15 here), when the digital input spike (Decay_start) from the TM adaptor is active (high). When there is no input spike, we will apply the stochastic approach (see equation (4)) to $V[t]$ in each time slot (of that TM exp-decay generator), until it reaches zero, indicating the end of the exponential decay.

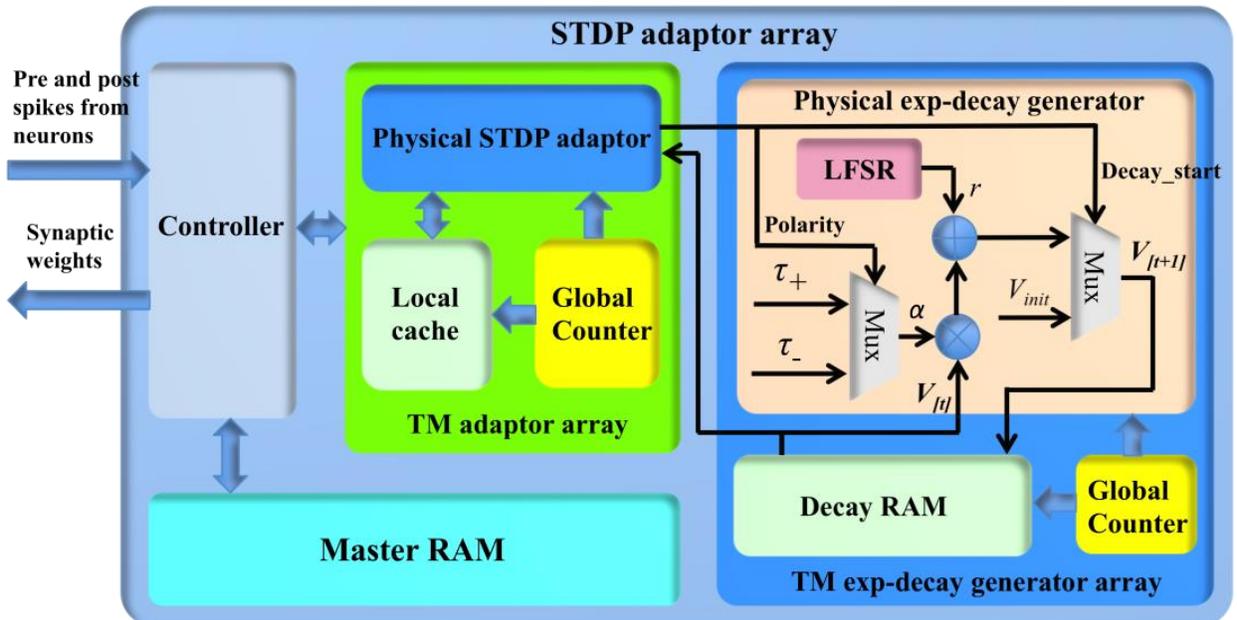

**Fig. 2. The structure of the STDP adaptor array.**

These computations were implemented with a single fixed-point-number multiplier. Its inputs are $\alpha$ (a 9-bit integer) and $V[t]$ (4-bit), resulting in a 13-bit output value. To allow different time constants for different synapses, we use a multiplexer to choose from different $\alpha$ s. Also for future extensions, we use a 7-bit LFSR to generate $r$, but the LFSR is configured to use only its five least significant bits in the work reported here. It will generate a new value every 1 ms. The integer part of the product, $V[t+1]$, will then be stored into the exp-decay RAM.

## IV. MEASUREMENT RESULTS

We have successfully implemented the exponential-type STDP adaptor array on an Altera Cyclone V FPGA. Table I shows the utilisation of hardware resources on the FPGA. As Table I shows, the proposed system uses only a small fraction (<1%) of the hardware resources.

We have tested the performance of the exponential-type STDP adaptor array by performing a balanced excitation experiment, based on the experiment run by [11]. Song et al.

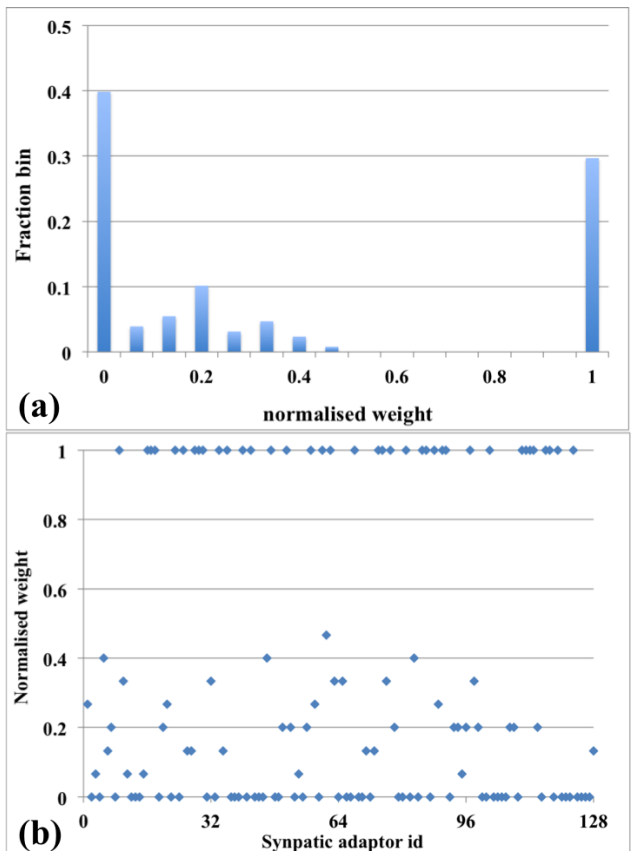

**Fig. 3. Balanced excitation experiment.** (a) Weight distribution after 1s of STDP for an input rate of 10 Hz. The bimodal distribution of *strong* and *weak* weights is apparent; (b) Scatter plot of the final weight distribution.

TABLE I

Device utilisation Altera Cyclone 5CGXFC5C6F27C7

| Layers | ALMs | RAMs | DSPs |
|--------|-----------|----------|--------|
| 1 | 246/29080 | 192k/4.5M | 1/450 |

have shown that competitive Hebbian learning [12] can be performed through STDP [11]. The competition (induced by STDP) between the synapses can establish a bimodal distribution of the synaptic weights: either towards zero (*weak*) or the maximum (*strong*) values (see Fig. 3).

## V. CONCLUSIONS

In this paper, we demonstrated a digital implementation of the STDP learning rule using a novel stochastic approach. This approach is capable of producing the same results to a more complex STDP model while occupying only a fraction of the area. The compactness of the implementation plus the added stochasticity of the results presents a perfect solution for implementing synaptic learning in large-scale digital neural networks.